\documentclass[a4paper,11pt,twocolumn]{article}

\usepackage{icphs2023}
\usepackage{metalogo} 
\usepackage{tipa}

\title{Prak: An automatic phonetic alignment tool for Czech}
\author{Václav Hanžl\textsuperscript{1}, Adléta Hanžlová\textsuperscript{2}}
\organization{\textsuperscript{1}Freelance researcher, \textsuperscript{2}Institute of Phonetics, Charles University in Prague}
\email{vhanzl@gmail.com, adleta.hanzlova@gmail.com}
\begin{document}

\maketitle

\begin{abstract}

Labeling speech down to the identity and time boundaries of phones is a labor-intensive part of phonetic research. To simplify this work, we created a free open-source tool generating phone sequences from Czech text and time-aligning them with audio.

Low architecture complexity makes the design approachable for students of phonetics.
Acoustic model ReLU NN with 56k weights was trained using PyTorch on small CommonVoice data. Alignment and variant selection decoder is implemented in Python with matrix library.

A Czech pronunciation generator is composed of simple rule-based blocks capturing the logic of the language where possible, allowing modification of transcription approach details.

Compared to tools used until now, data preparation efficiency improved, the tool is usable on Mac, Linux and Windows in Praat GUI or command line, achieves mostly correct pronunciation variant choice including glottal stop detection, algorithmically captures most of Czech assimilation logic and is both didactic and practical.

\end{abstract}

\keywords{phonetic alignment, segmentation, PyTorch, Czech, Praat}

\section{Introduction}
Labeling speech recordings and identifying phone boundaries is a significant part of phonetic research. Even though there are software tools to automate this process, many of them target only languages with great quantities of speakers and most are also not freely available, their use requires complex installation and can be restricted by license agreements. Tool choices are even more limited for less common languages like Czech.

Praat's \cite{praat} own integrated alignment tool is usable when aligning individual words or short sentences, but its performance is significantly worsened when the audio contains pauses. It also only works from the Praat \textit{View \& Edit} window menu and doesn't enable automatic alignment of larger datasets.

The Czech aligner used in academic research so far and preferred as the most useful available 
is Prague Labeller  \cite{pollak2005hmm, pollak2007hmm} which is based on HTK GMM models and shows admirable longevity.
Subsequent research used Kaldi \cite{SegmentationKaldiCzech}.
Other ad-hoc alignment solutions were never finalized as universally usable.

As identifying words and phones and their time boundaries in a recording is very useful in phonetics not only for research purposes, but also for educational needs and general better orientation in the audio, we find it important that a cross-platform tool enabling the automation of such processes and striving to be as precise as possible is widely available and easy to access by phoneticians and students and requires no programming knowledge from the user. In this paper, we present Prak -- a tool we developed with this idea in mind.

\section{Design goals and scope}
Observing current practices when preparing Czech phone-aligned research data, we had several goals for our new alignment tool:
\begin{itemize}
\item an open-source tool free for any use -- MIT license \cite{MITlicense} for code, trained on free audio data
\item functioning on Mac, Linux and Windows
\item easy to install -- low dependencies, only reliable\break
 dependencies~which~are~(hopefully)~here~to~stay
\item simple architecture, preferably building on techniques from phonetics students' curricula
\item usable from both GUI and command line
\item using explainable and modifiable logic (rather than a trained blackbox) where possible
\item automatic pronunciation variant selection
\end{itemize}

The tool expects an audio recording and text transcript as an input. While we can imagine doing the transcript via ASR, we deliberately left this task to an external tool if so desired (manual transcription is a noticeably smaller task than phone boundary adjustments). This way we can always use a state of the art ASR and combine it with our phone alignment tool, without modifying the ASR tool for phone alignment needs -- merging these tasks into one tool used to be a vital approach in the GMM HTK era, but modern ASR tool construction often abandons the notion of phones altogether.

\begin{figure}[!ht]
\begin{center}
\setlength{\fboxsep}{0pt}
\fbox{\includegraphics[width=8cm]{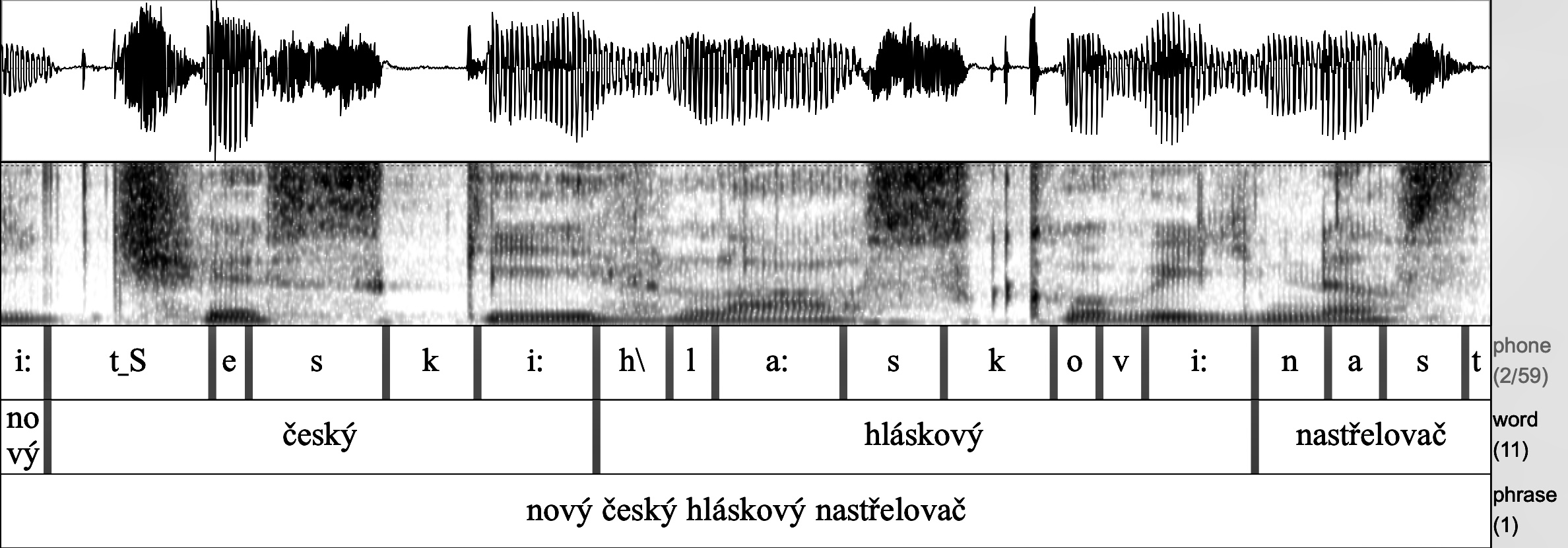}}
\caption{Output textgrid of Prak used in Praat.}\label{fig:nastrel}
\end{center}
\end{figure}

The initial release supports Czech language only, as practical help to Czech research was our primary goal. Nevertheless, we tried to make extensions to more languages easy by using a multilingual training data source and by modular design of the code.

\section{Design}
\subsection{Training data}
Czech language has high quality speech corpora for acoustic model (AM) training, however, many data sets are unsuitable for a free tool.
Some are commercial and costly (SpeechDAT \cite{SpeechDat}, Speecon \cite{Speecon}), some free of cost, but for research use only.
There is no LibriSpeech \cite{LibriSpeech} or TEDx for Czech.
Corpus of audio recordings from the Chamber of Deputies of the Parliament of the Czech Republic available freely online \cite{ParCzech3, nahrParlament} is big and free, but contains reverberations stronger than standard studio recordings. Some Czech TV recordings are available, but have a low number of speakers.

We decided to first try the CommonVoice (CV) \cite{CommonVoice} dataset, the size of which is marginal for Czech (14 hours of training data, 4 hours being silence), but it has suitable recording conditions, contains a high number of speakers, has an extremely permissive license and is available for many other languages. To our surprise, CV itself allowed us to train an AM of sufficient quality, so we initially released Prak with a model trained on CV alone.

We are also very grateful to the Institute of Phonetics, Charles University in Prague, for providing their manually labeled recordings from which we selected 5 hours of easily usable data. We did not use this data in any training to keep our tool free of additional dependencies, but we used 10\% of this dataset to evaluate the tool by comparison to human labeling made by potential users themselves.

\subsection{Software tools}

\begin{figure}[!ht]
\begin{center}
\includegraphics[width=7.5cm]{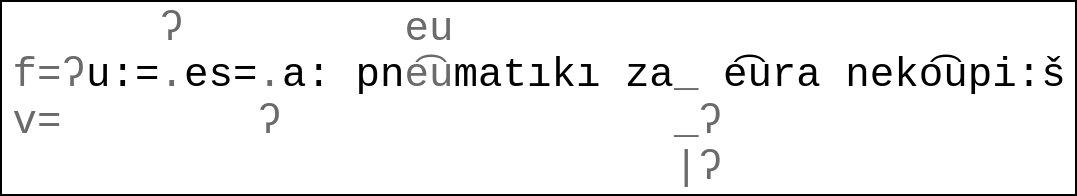}
\caption{Command line output showing phonetic transcription with pronunciation variants.}\label{fig:pronpneu}
\end{center}
\end{figure}

Currently available free and precise deep learning ASR tools \cite{Whisper} naturally come to mind as building blocks, however, the shift towards direct mapping between text and audio omits the phone level, requiring significant modifications of these complex tools to repurpose them for phone alignment.

From older ASR tools, Kaldi \cite{Kaldi} is still close to the traditional HMM-based approach of HTK and explicitly uses phones. It is however quite a complex dependency in a software system. We therefore used a modern NN toolkit PyTorch \cite{PyTorch} for construction of the phone AM, but designed our ASR-like infrastructure from scratch in Python \cite{python3} using a classical HMM approach. As phone alignment is quite a limited subset of ASR, we expected this to be feasible.

The matrix library in PyTorch also has a very close alternative in the even more broadly used NumPy \cite{NumPy}, allowing easy reimplementation of at least the inference part, should such need ever arise.

\subsection{Integration in Praat GUI}
The Praat \cite{praat} integration of Prak is designed to be simple to use without any programming knowledge. A Praat script which embeds the main Python tool is easy to add to Praat dynamic menu, so users can align text with audio in just one click -- see fig. \ref{fig:nastrel} for example result. The script also performs several input file checks and adds additional features such as aligning multiple sounds using one text source.

\section{Pronunciation generator}

\subsection{Phonetic alphabets}
Alignment output in Praat TextGrid files uses SAMPA \cite{CzechSAMPA} to stay compatible with established practice. Terminal (command line) output uses much more readable 
IPA for Czech as used in \cite{foneticka_segmentace_2010}, see fig.~\ref{fig:pronpneu} for an example of generated pronunciation and its variants. Internally (in the source code) the tool uses programming-friendly encoding where each phone is represented by one Czech character.

\subsection{Text cleanup}

\begin{figure*}[!ht]
\begin{center}
\includegraphics[width=17cm]{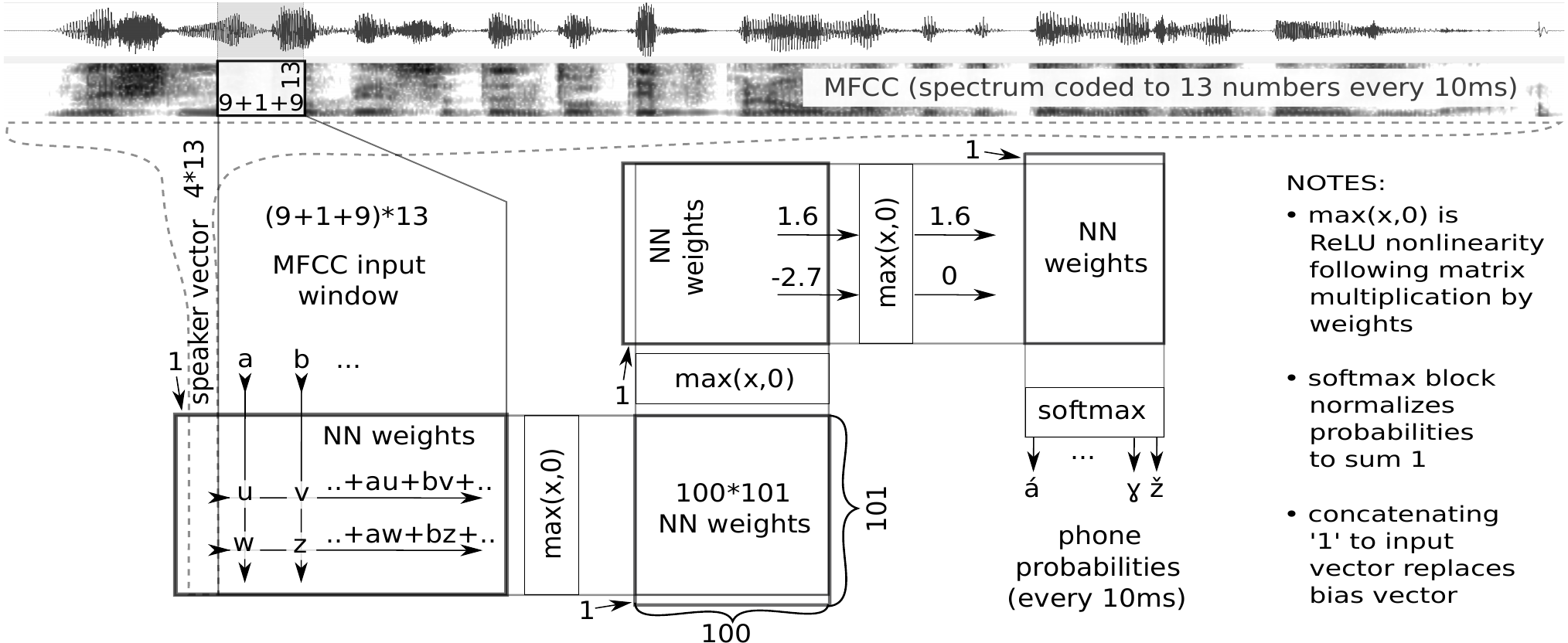}
\caption{Acoustic model with ReLU NN stack converting MFCC window to phone probabilities.}\label{fig:AM_NN}
\end{center}
\end{figure*}

Great care was taken to be able to process all usual texts
on all the supported platforms. Apart from notoriously known end-of-line encoding differences \cite{CRLFissues}, there are more subtle pitfalls like invisible Unicode BOM characters, NFC and NFD forms of accented characters etc. We hopefully made it safe to process all the variants happening in practice.

\subsection{Foreign words}
Users can add exceptions in the form of replacement rules. For each word, the longest match in rules is replaced by all pronunciation variants from the rule (e.g. rule "washington vošingtn" can do replacement in "Washingtonu"). Replaced text is not matched anymore, but parts of words before and after replacement are subject to potential additional (shorter) matches. The resulting text eliminates differences~between~foreign~and~native~Czech~words.

We find the longest match priority more practical than the approach used in \cite{pollak2007hmm} where rules are tried in sequence, making it hard to identify the right place for adding new rules among hundreds of old ones.

\subsection{Assimilation logic in backward FSTs}
Many Czech assimilations can be expressed as Finite State Transducer (FST) with a small number of states (often just 2 or 3) processing the phone sequence backwards, including rather far distance effects. For example, in the sequence "vošingtnu" (made by the replacement rule above), processing [\textipa{t}] changes FST state to \textit{devoicing} and [\textipa{g}] turns into voiceless [\textipa{k}]. In another FST, [\textipa{k}] or [\textipa{g}] forces state \textit{velarization} and affected [\textipa{n}] becomes [\textipa{N}]. We use a convolution of FSTs taking care of:
\begin{itemize}
\item  glottal stop [\textglotstop], intervocalic [\textipa{j}]
\item  assimilation of voicing
\item  consonant groups containing dtn/ďťň such as “ntní” [nt\textltailn{}i\textipa{:}/nc\textltailn{}i\textipa{:}/\textltailn{}c\textltailn{}i\textipa{:}]
\item  bě/pě/vě/mě/fě [\textipa{bjE, pjE, vjE, m\textltailn{}E, fjE}]
\item  velarization in nk/ng [\textipa{Nk/Ng}]
\end{itemize}
FSTs provide multiple outputs (needed esp. on word boundaries), leaving the final choice to the AM.

\section{Phone acoustic model}


While transformers \cite{Transformer, TransformerSurvey} of various kinds or at least convolutional structures come to mind as appropriate state of the art in speech processing, our limited goal of only phone alignment seemed achievable with much simpler architectures, significantly~lowering~the~barrier~for~students~trying to grasp the inner workings of the tool. We therefore started with a simple stack of layers composed of matrix multiplication and ReLU -- Rectifying Linear Units, $\max(x,0)$ -- applied to each element of a vector~which~is~passed~between~layers~(see~fig.~\ref{fig:AM_NN}).

As input, the NN gets 19 consecutive MFCC frames plus a "speaker vector" which is an average MFCC in 4 energy bands (frames are split to above and below average energy and then sub-split again).
We compute Kaldi-like MFCCs via PyTorch.

The stack of ReLU layers only maps one position in the audio to one phone hypothesis, not sequence to sequence. For this purpose, we stuck to a classical HMM structure, replacing a GMM model with ReLU NN. Our training is a cross between Baum-Welch reestimation and NN training by gradient descent, alternating phases of time re-alignment of the target phone sequence and gradient descent of the AM guessing these phones.

\begin{figure}[!ht]
\begin{center}
\includegraphics[width=8cm]{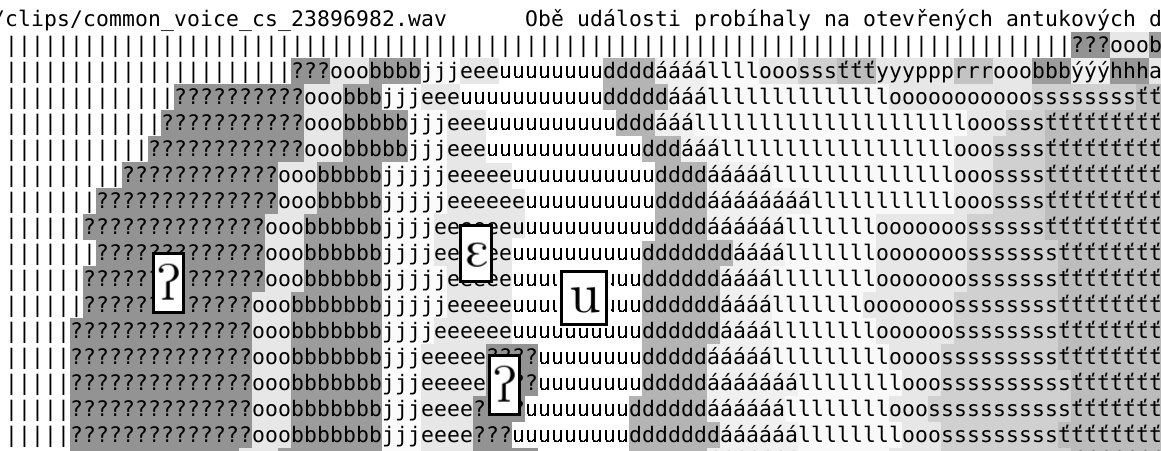}
\caption{Phone alignments during training, rows show train epochs. An optional second glottal stop is only found later, leading to
[\textglotstop\textipa{objE}\textglotstop\textipa{u}...]}\label{fig:train}
\end{center}
\end{figure}

Phone~sequences~are~guessed~by~Viterbi~alignment using phone probabilities estimated by NN AM, while the NN AM itself is trained by gradient descent, using the phone sequences as training targets.
Tools like HTK solve similar circular dependencies in Baum-Welch reestimation by a wild initial alignment guess. What we do is similar: For the initial alignment, every phone is presupposed to be 30ms long with equal silences preceding and following speech in a recording. The system is able to converge from this (mostly false) initial alignment to a very good phone AM and very good phone alignments.~Fig.~\ref{fig:train}~shows~example~training~dynamic.

The only adjustment needed for convergence is an artificial boost of rare phones. In Viterbi decoding, we adjusted phone probabilities according to the current overall number of frames classified as this phone in the whole training dataset.

\section{Evaluation of phone alignments}
We compared the performance of Prak with Prague Labeller \cite{pollak2007hmm}, using manually aligned data as a reference. Direct comparison is not easy as phone sets differ between aligners (e.g. Prak also detects glottal stops, uses both voiced and voiceless "ř", considers assimilation at word boundaries etc.). The manual alignment, while made with great effort and care, uses several slightly different approaches and contains some phone identity errors.
An important parameter for practical users is the frequency of misalignments which must be corrected by moving multiple phone boundaries, therefore we tried to detect major phone boundary misplacements.

We compared pronunciations, counting phone insertions, deletions and substitutions. At places of match, we measured phone center time shifts against the manual reference (while the exact phone identity may be questioned,\footnote{We manually checked the test set and corrected obvious errors, however given its size (there are 20k phones in the 10\% test subset of the full 5 hour set), we certainly did not correct all errors even by this additional pass of human work.}
the time positions are very exact in our manual reference data). The count of time mismatches exceeding a threshold was used as a quality measure, results are shown in tab. \ref{tab:hardmis}.

\begin{table}[!ht]
\begin{center}
\begin{tabular}{|c|c|c|}
\hline
\rowcolor[gray]{.75}
test   & PL \cite{pollak2007hmm}    & Prak\\
\hline
phone mismatch             & \kern5.3pt$6.61$  &   $1.88$\\
match but misplace 0.1s+   & \kern5.3pt$\mathbf{4.28}$  & $\mathbf{0.36}$\\
match but misplace 0.2s+   & \kern5.3pt$3.22$  &   $0.09$\\
mismatch or misplace 0.1s+ &   $10.89$  &   $2.24$ \\
\hline
\end{tabular}
\caption{Percentage of phone mismatch and boundary misplacement, comparing most used tool to ours.}\label{tab:hardmis}
\end{center}
\end{table}

\section{Future work}
We~highly~emphasized~simplicity,~the~only~bigger
tool~being~a~deep~learning~framework,~only~applied
to~a~basic~ReLU~stack.~We~avoided~FST~toolkit~(a
basic~tool~which~Kaldi~explicitly~uses,~unlike~HTK
where~this~functionality~is~present~but~dissolved~in
higher~layers)~and~used~a~simple~"sausage"~structure
to~capture~pronunciation~variants.~There~are~many
wonderful~ready-to-use~building~blocks~in~current
ASR~toolkits~which~would~be~great~to~test~in~this~task.

While~modern~ASR~tools~abandoned~explicit
design~of~middle~layer~representation~like~phonemes
or~phones,~it~would~be~possible~to~map~text~to~a~phone
sequence~and~train~e.g.~wav2vec2~\cite{wav2vec2}~architecture
on~phones~instead~of~characters~and~extract~time
boundaries~of~phones~using~a~method~such~as~\cite{motohira}.

Another option would be to use more complex NN structures like transformers for the AM.
The question is how much would time boundaries be smeared by the structure paying attention to too distant parts of audio. Such AM would perhaps only be useful as a first layer alignment anchoring audio to text on the level of words, followed by fine-grained alignment by a more local AM like ours.

Phone boundaries could be further fine-tuned.
Movements of boundaries towards spectrum change might help for some phone pairs, different processing for others -- in fact, the challenge here is translating \cite{foneticka_segmentace_2010} into a programming language.

Our speaker-vectors can certainly be replaced by i-vectors \cite{Kaldi} or x-vectors \cite{xvectors}, trading better results for considerably increased complexity.

We hope that our alignment tool can serve as a useful framework and baseline for such experiments.

\section{Conclusion}
We~made~a~practical~Czech~phone~alignment
tool~which~significantly~outperforms~the~best~tool
available~so~far~in~the~boundary~misalignment
measure.~Prak~also~provides~new~features~like
automatic~variant~selection~and~is~more~resilient~to
problematic~input~texts.~We~consider~our~biggest
achievement~that~we~gave~researchers~and~students
an~open-source~tool~with~no~restrictions~and~simple
to~understand~architecture,~hopefully~allowing~others
to~build~on~it.~Free~download~of~the~source~code~is
possible~here:~{\tt~github.com/vaclavhanzl/prak}

\bibliographystyle{IEEEtran}
\bibliography{icphs2023}

\theendnotes

\end{document}